\documentclass{article}
\usepackage{spconf,amsmath,graphicx}
\usepackage{subfig}
\usepackage{url}

\title{Localization of Ice-Rink for Broadcast Hockey Videos}
\name{Mehrnaz Fani, Pascale Brunelle Walters,  David A. Clausi, John Zelek and Alexander Wong}
\address{Systems Design Engineering, University of Waterloo}
\begin{document}
%
\maketitle
%
\begin{abstract}
   In this work, an automatic and simple framework for hockey ice-rink localization from broadcast videos is introduced. First, video is broken into video-shots by a hierarchical partitioning of the video frames, and thresholding based on their histograms. To localize the frames on the ice-rink model, a ResNet18-based regressor is implemented and trained, which regresses to four control points on the model in a frame-by-frame fashion. This leads to the projection jittering problem in the video. To overcome this, in the inference phase, the trajectory of the control points on the ice-rink model are smoothed, for all the consecutive frames of a given video-shot, by convolving a Hann window with the achieved coordinates. Finally, the smoothed homography matrix is computed by using the direct linear transform on the four pairs of corresponding points. A hockey dataset for training and testing the regressor is gathered. The results show success of this simple and comprehensive procedure for localizing the hockey ice-rink and addressing the problem of jittering without affecting the accuracy of homography estimation. 
\end{abstract}
\section{Introduction}

In computer vision based sports video analytics, one of the fundamental challenges is precise localization of the sport field. Which is determining where the broadcast camera is looking, at any video frame. Result of this can be used for different purposes, such as determining the players’ coordinates on the sport field, camera calibration \cite{chen2019sports}, placement of the visual ads on the sport field. 

In the literature, this problem is addressed by finding the homography between video frames and the top-view model of the sports field. Generally, the state-of-the-art procedures for localizing the sports fields \cite{citraro2020realtime, jiang2020optimizing, sha2020endtoend} 
utilize deep structures for registering the video frames on the model of the sport filed, by minimizing a distance metric between the ground-truth and predicted points (or areas) of a frame on the model (or vice versa). These predictions are usually performed in a frame-by-frame manner, so, the result of registration (localization) for a given video sequence usually suffers from jittering. Therefore, an additional smoothing step on predicted homography of the consecutive frames is required. For smoothing or refinement of the initial homography estimation, some 
works in the literature such as\cite{chen2019sports}, use Lucas-Kanade algorithm \cite{baker2004lucas}, which is based on tracking feature patches in the video frames. So, the success of this procedure highly depends on the detecting precision of the corresponding features.

In this work, a shot boundary detection procedure for broadcast hockey videos is implemented by adopting the cut transition method in \cite{yazdi2016shot}. A frame-based hockey ice-rink localization is performed by inspiration from \cite{jiang2020optimizing}. A ResNet18-based regressor is implemented which, for an input frame, regresses to the four control points on the hockey ice-rink model. In the inference phase, to remove the jittering of warped frames on the ice-rink model, for consecutive video-frames, the trajectories of the control points on the ice-rink model are smoothed by using a simple moving average operation based on Hann window. Then, the homography matrix is computed by using the direct linear transform (DLT) method \cite{dubrofsky2008combining}.
To the best of our knowledge there is no publicly available dataset on hockey videos for ice-rink localization. Therefore, a dataset of 72 hockey video-shots, along with their key-frames and their corresponding homography matrices, is generated and used to train, evaluate and test our method.

Our contributions in this paper are:
\begin{itemize}
    \item Adopting a cut transition detection procedure, for fast and accurate shot boundary detection in broadcast hockey videos.
    \item Implementing a frame-by-frame regressor for localizing the video-shot frames on the ice-rink model.
    \item Introducing a new smoothing procedure by convolving a Hann window to the trajectory of the predicted control points, for the successive video-shot frames.
    \item Generating a new hockey data set for ice-rink localization.
\end{itemize}

\section{Related Works to Sports Field Localization}

In this section, the related works in the literature for localizing the sports fields are reviewed.

Sports field localization is a special case of homography estimation, where the structure of the playing field plane is known \cite{nie2021robust}. 

Homayounfar \textit{et al.} detect lines with a VGG16 semantic segmentation network, use them to minimize the energy of the vanishing point, and estimate the camera position via branch and bound \cite{homayounfar2017sports}. 
Sharma \textit{et al.} and Chen and Little use the pix2pix network to extract the lines from the playing surface on a dataset of soccer broadcast video. Both works then compare the extracted edge images to a database of synthetic edge images with known homographies in order to localize the playing field \cite{chen2019sports, sharma2018automated}.

Cuevas \textit{et al.} detect the lines on a soccer field and classify them to match them to a template of the field \cite{cuevas2020automatic}. Tsurusaki \textit{et al.} use the line segment detector to find intersections of lines on a soccer field, then match them to a template of a standard soccer field using an intersection refinement algorithm \cite{tsurusaki2021sports}.

Sha \textit{et al.} detect the zones from soccer and basketball datasets. They initialize the camera pose estimation through a dictionary lookup and refine the pose with a spatial transformer network \cite{sha2020endtoend}. 

Tarashima performs semantic segmentation of the zones as part of a multi-task learning approach for a basketball dataset \cite{tarashima2020sflnet}. 
Citraro \textit{et al.} segment keypoints based on intersections of the lines on the playing surface and match them to a template for basketball, volleyball, and soccer datasets \cite{citraro2020realtime}.
Similarly, Nie \textit{et al.} segment a uniform grid of points on the playing surface and compute dense features for localizing video of soccer, football, hockey, basketball, and tennis \cite{nie2021robust}.

Jiang \textit{et al.} propose a method for refining homography estimates by concatenating the warped template and frame and minimizing estimation error. They report results on soccer and hockey videos \cite{jiang2020optimizing}.

The wide variety of methods that have been described in the literature shows that there is no one method that works particularly well for all sports applications. Recently described techniques show that deep network architectures achieve better performance with faster computation \cite{jiang2020optimizing, tarashima2020sflnet, citraro2020realtime, chen2019sports}. 

\section{Methodology}

\begin{figure*}[t]
	\begin{center}
		\subfloat(a) SBD{
		    \includegraphics[width=0.83\linewidth]{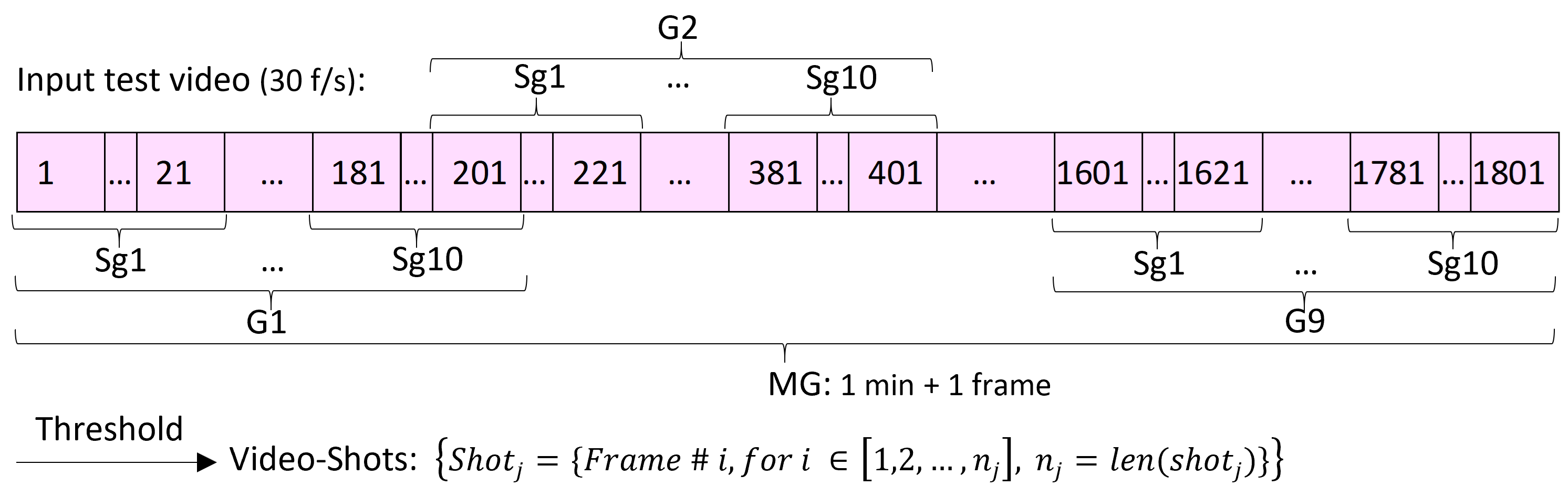}
		}\\
	
		\subfloat(b) Regressor{
		\includegraphics[width=0.8\linewidth]{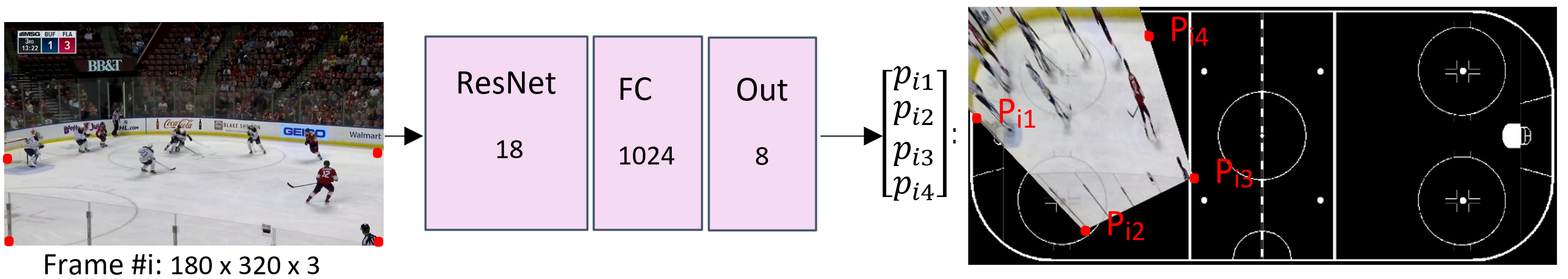}
		}\\
	
		\subfloat(c) Smoothing{
		\includegraphics[width=0.8\linewidth]{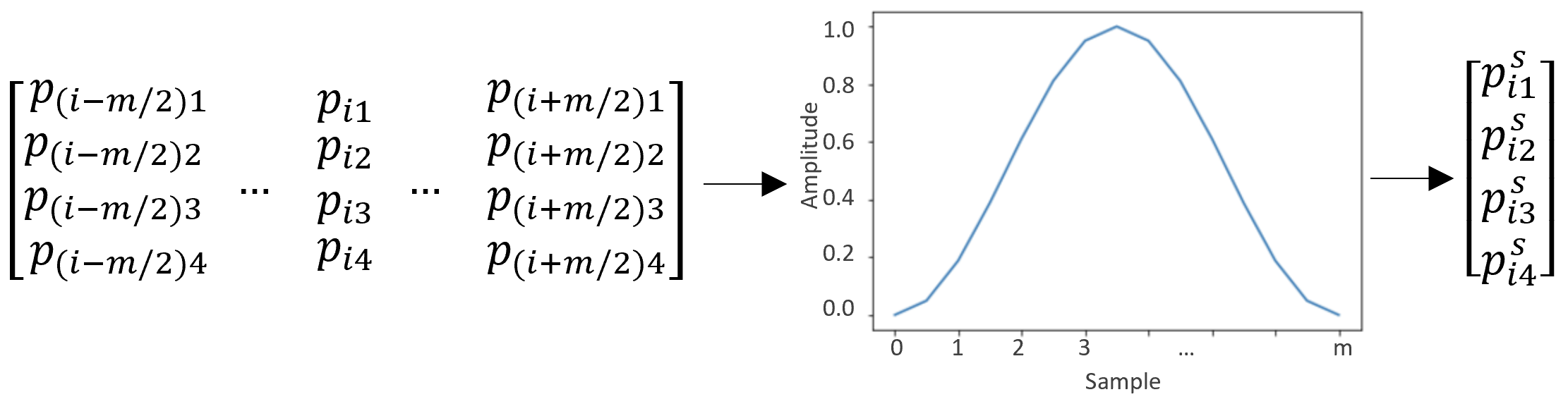}
		}		
	\end{center}
	\caption{General framework for localizing the hockey video-shots on the ice-rink model. (a) Shot boundary detection is performed to segment the input video into smaller temporal units, i.e., video-shots. (b) A ResNet 18-based regressor is implemented to regress to the four control points, for all frames of the input video-shot (c) By using a moving average operation, trajectory of the control point on the ice-rink model is smoothed for the input video-shot.}
	\label{fig:framework}
\end{figure*}

The implemented method to localize hockey ice-rink for an input hockey video has three main steps, which are illustrated in Fig.\ref{fig:framework} and are listed here: 1) Shot boundary detection (SBD), which is breaking the input video into smaller temporal sequences, called video-shots\cite{yazdi2016shot}, 2) Localizing each frame of the video-shot on the top-view model of the ice-rink, by regressing into four control points on the model\cite{jiang2020optimizing}. 3) Smoothing the trajectory of the control points for the input video-shot, using a moving window, and calculating the smoothed homography matrix for each frame by using the DLT algorithm \cite{dubrofsky2008combining}. These steps are explained thoroughly in the following subsections.

\subsection{Shot Boundary Detection}
\label{sec:SBD}
Shot boundary detection (SBD) is the primary step for any further video analysis, that temporally breaks a video into smaller units, called video-shots. Each video-shot is composed of a sequences of frames that are semantically consecutive and are  captured with a single run of a camera. Here, SBD is performed by adopting the procedure used for cut transition detection in \cite{yazdi2016shot}. As shown in \ref{fig:framework}(a), first, the input video with frame-rate of 30 fps, is hierarchically partitioned. One-minute window (plus one extra frame) of the video (i.e., 1801 frames) is considered as a mega-group (MG). Each mega-group includes nine groups (G), and each group includes ten segments (Sg), where each segment includes 21 frames. The neighbouring segments within a group have one frame overlap. The neighbouring mega-groups have one segment overlap. Based on this partitioning a thresholding mechanism is defined to detect the video-shot boundaries. \\
Each frame is divided to 16 blocks. The block histogram of the marginal frames (first and last frame) of the segment are computed. If the distance of the marginal histograms for a segment goes beyond the group threshold, $T_G = 0.5 \mu_{G} + 0.5 \left[ 1+\ln\left(\frac{\mu_{MG}}{\mu_{G}}\right)\sigma_{G} \right]$, that segment would be a candidate for including shot transition.\\
For each candidate segment, the differences between block histograms of all the successive frames are computed. If the maximum difference in a segment goes beyond a threshold, i.e., $\frac{max(dist_{sg})}{\mu_{G}} > \ln\left(\frac{\mu_{G}}{\mu_{Sg}\sigma_{Sg}}\right)$, the frame associated with the maximum difference is detected as the shot boundary. 

\subsection{Localizing Frames on the Ice-Rink Model}

Localizing hockey frames on the ice-rink model is performed in a frame-by-frame fashion by using a ResNet18-based regressor which is shown in Fig.~\ref{fig:framework}(b). The ResNet18 model is pre-trained on ImageNet dataset and two fully connected layers are appended to the network.\\

\textbf{Training.} The network is trained on our dataset of NHL hockey video key-frames annotated with ground truth homographies. For each key-frame of size $h\times w \times 3$, four pre-determined control points are considered, i.e., $[[0, 0.6h], [0, h],$\\ $[w, h], [w, 0.6h]]$. Having the ground truth homography of the key-frames, the network is trained to regresses the four projected control points on the ice-rink model, by minimizing the squared error between the ground-truth points, $P_{gt}$, and the estimated points, $P_{est}$ as per eq. \ref{eq:loss}.

\begin{equation}
\label{eq:loss}
    L =\| P_{est}-P_{gt}\|^2_{2}
\end{equation}

\textbf{Inference.} In the inference phase, each frame of the input video-shot, i.e., $Frame_{i}, i \in \{1,2,...,len(shot)\}$, is fed into t he network and the four control points on the ice-rink model, i.e., $[p_{i1}, p_{i2}, p_{i3}, p_{i4}]$, are estimated. Having the four pre-determined control points on the frame, the four pairs of points are used to calcule the homography matrix by using the DLT algorithm. However, after warping the successive frames of the video-shot on the ice-rink model according to the inferred homography matrices, a jittering phenomenon is observed. This is due to the frame-by-frame homography estimation method. Incorporating the temporal information in calculation of the homography matrix, as explained in the next sub-section, can smooth the projection of the video-shot.       

\subsection{Smoothing the Trajectories of Points }

Here, a straightforward, and effective smoothing method has been proposed and implemented, which is based on smoothing the trajectory of the four estimated control points on the ice-rink model.\\

\textbf{Hypothesis.} For a given video-shot we can assume that camera moves smoothly in every direction in the 3D space. In other words, camera's field of view (FOV) changes smoothly in the 3D space. A mapping of the 3D-FOV of the camera on the 2D space can be achieved by homography. Therefore, for a video-shot, changes of the 2D-FOV of the camera (i.e., changes of the consecutive warped frames on the ice-rink model) should be smooth as well. From this, we can directly conclude that the four control points should have an smooth trajectory on the ice-rink model. \\

A Hann window of size $m+1$ (with an odd size) is used to smooth the trajectory of the control points on the ice-rink model. The coefficient of this window can be computed as given in eq. \ref{eq:Hann}.

\begin{equation}
    \label{eq:Hann}
    w(n)= 0.5 \left( 1-cos(2\pi\frac{n}{m})\right)
\end{equation}

Fig.~\ref{fig:framework}(c) demonstrate how the estimated control points of $Frame~\#i$ are being smoothed by applying the Hann window on control points of $m+1$ neighboring frames, i.e., frames $\frac{(i-m)}{2}$  to  $\frac{(i+m)}{2}$, which is essentially a convolution operation. The smoothed homography is then computed by using DLT method between the smoothed points, $[p_{i1}^{s}, p_{i2}^{s}, p_{i3}^{s}, p_{i4}^{s}]$, and the four fixed control points on the frame.

\section{Results}
In the following subsections some quantitative an qualitative experimental results are given. As there is no publicly available dataset of hockey, for ice-rink localization purpose, to conduct the experiments, we have prepared a dataset that is described here.

\textbf{Dataset.} Our dataset includes 72 video-shots (30 fps), that are captured from 24 NHL hockey games, by using the SBD procedure that is explained in section \ref{sec:SBD}. For each video-shot, a number of key-frames are extracted sparsely (i.e., 20 frames apart from each other) and are annotated to get their ground truth homography matrices. Annotations are collected by annotating point correspondences on each hockey broadcast frame and the rink model. Key-frames from 11 video-shots of four games are used for validation, the key-frames from 11 video-shots of four other games are used for test and rest of the data is used for training. Data augmentations, such as flipping the frames, are performed on the training data. Inference is performed on all frames of test video-shots.

\subsection{Qualitative Evaluation of Smoothing}

\begin{figure}[!t]
	\begin{center}
		\subfloat{
		    \includegraphics[width=1\linewidth]{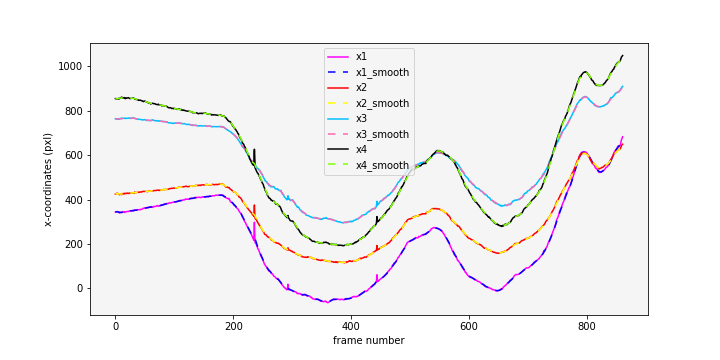}(a)
		}\\
	
		\subfloat{
		\includegraphics[width=1\linewidth]{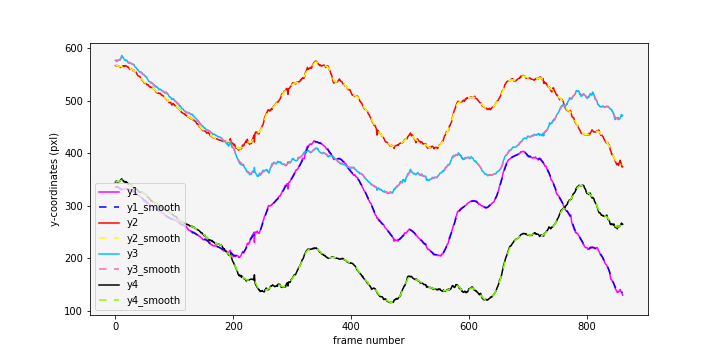}(b)
		}\\
	
		\subfloat{
		\includegraphics[width=1\linewidth]{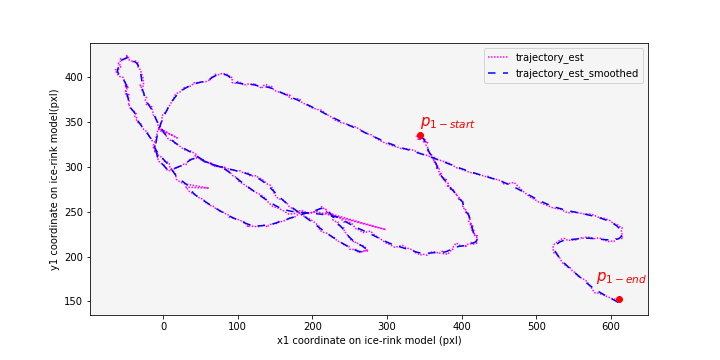}(c)
		}		
	\end{center}
	\caption{Smoothing the trajectory of the four control points for a test video-shot using the Hann window reduces jittering in the output. (a) The x-coordinates of the four control points in successive frames of the video-shot, before and after smoothing. (b) The y-coordinates of the four control points in successive frames of the video-shot, before and after smoothing. (c) The trajectory of control point $p_{1}$ on the ice-rink model coordinates, before and after smoothing.}
	\label{fig:smoothing}
\end{figure}

The effect of smoothing the trajectory of the four control points for the consecutive frames of a test video-shot is demonstrated in Fig. \ref{fig:smoothing}. The x-coordinates and the y-coordinates of the four points, i.e., $ [p_{1}, p_{2}, p_{3}, p_{4}] = [(x_{1}, y_{1}), (x_{2}, y_{2}), (x_{3}, y_{3}), (x_{4}, y_{4}) ]$, are smoothed separately by using the Hann window of size 15. The x and y-coordinates, before and after smoothing, are coded with different colors and are shown in Fig. \ref{fig:smoothing}(a) and Fig. \ref{fig:smoothing}(b). Also the trajectory of point $p_{1}$ on the ice-rink, before and after smoothing, is given in Fig. \ref{fig:smoothing}(c). 
These figures show the jittering phenomenon (over and under shooting of the estimated coordinates) in the trajectories of the control points. The smoothing removes this jittering noise. In practice, when the ice-rink model is projected on the frames using the smoothed homography, the resulting video-shot is visually continuous and free from jittering.

\subsection{Quantitative Evaluation of Smoothing}

The effect of smoothing on IOU\textsubscript{part} as well as the accuracy of projected points are evaluated and given in Table \ref{tab:accuracy}. Our method before and after applying smoothing step is compared to single feed forward (SFF) network \cite{jiang2020optimizing}. The result for SFF is directly reported from the related paper. The experiment shows that the results almost remain the same with or without applying the homography smoothing.

In the second experiment a smoothness factor for the test video-shots are computed and reported in Table \ref{tab:smoothing_eval}.
The network regresses the positions of four control points on the broadcast frame after it has been warped onto the rink model. 
The smoothness score is the average of the smoothness of the four control points. Smoothness is calculated for a sequence of points by calculating the standard deviation of the differences of each of the coordinates. The score is normalized by the mean difference. For this score, lower is better. Smoothing the network output gives a lower mean smoothness than directly using the network smoothness.

\begin{table}[!t]
    \centering
    \caption{Mean IOU\textsubscript{part}, mean, median and variance of projected points on the ice-rink model, for our method before and after smoothing (BFS and AFS). A comparison with SFF \cite{jiang2020optimizing} is also performed.}
    \footnotesize
    \setlength{\tabcolsep}{0.2cm}
    \begin{tabular}{|c|c|c|c|c|}\hline
       method & mean IOU\textsubscript{part} & Mean\textsubscript{proj} & Var\textsubscript{proj} & Median\textsubscript{proj} \\\hline\hline
       Ours(BFS)& 94.63 & 12.87 & 53.22 & 11.64 \\
       \hline
       Ours(AFS) & 94.65 & 12.91 & 54.02& 11.90\\
       \hline
       SFF\cite{jiang2020optimizing} & 90.1 & - & -&-\\
       \hline
    \end{tabular}
    \label{tab:accuracy}
\end{table}

\begin{table}[!t]
    \centering
    \caption{Smoothness evaluation of the hockey ice-rink localization before and after smoothing step ( i.e, S\textsubscript{BFS} and S\textsubscript{AFS}), and the difference of the two are reported in this table.}
    \footnotesize
    \setlength{\tabcolsep}{0.2cm}
    \begin{tabular}{|c|c|c|c|c|}\hline
        Test video-shots & S\textsubscript{BFS} & S\textsubscript{AFS} & Difference \\
        \hline\hline
        $Video-Shot_{1}$ & 33.62 & \textbf{14.20} & 19.42 \\
        \hline
        $Video-Shot_{2}$ & 131.90 & \textbf{97.96}& 33.93 \\
        \hline
        $Video-Shot_{3}$& 93.64 & \textbf{34.31} & 59.34 \\
        \hline
        $Video-Shot_{4}$ & 221.50 & \textbf{201.81} & 19.68 \\
        \hline
        $Video-Shot_{5}$ & 20.96 & \textbf{12.88} & 8.08 \\
        \hline
        $Video-Shot_{6}$ & 59.98 & \textbf{49.51} & 10.46 \\
        \hline
        $Video-Shot_{7}$ & 34.94 & \textbf{13.26} & 21.68 \\
        \hline
        $Video-Shot_{8}$ & 183.80 & \textbf{124.93} & 58.87 \\ 
        \hline
        $Video-Shot_{9}$ & 290.31 & \textbf{262.14} & 28.16 \\ 
        \hline
        $Video-Shot_{10}$ & 58.49 & \textbf{40.17} & 18.31 \\
        \hline
        $Video-Shot_{11}$ & 411.07 & \textbf{327.46} & 83.60 \\
        \hline
    \end{tabular}
    \label{tab:smoothing_eval}
\end{table}

The mean output smoothness before smoothing (S\textsubscript{BFS}) across all test video shots is 140.02, and after smoothing (S\textsubscript{AFS}) is 107.15. This is a reduction of 32.86, meaning that the smoothing step increases the measured smoothness.

\section{Conclusion}

In this work, hockey ice-rink localization from broadcast videos is performed. The major focus of our work was finding an efficient framework for breaking the raw hockey videos into shots, finding the homography of each video-shot, and addressing the jittering phenomenon that occurs due to the frame-by-frame nature of estimating the homography. To fulfill our aims, we first adopted a threshold-based shot boundary detection approach and applied that on hockey videos. Then, in order to calculate the homography between each frame of the shot and the ice-rink model, like similar works in the literature, we implemented a ResNet18-based regressor to regress into four control points of the frames on the model. In our last stage, we smoothed the predicted control points of a frame by convolving a Hann window on the trajectory of the control points in $m+1$ neighboring frames. The achieved results show that smoothing does not improve or worsen the accuracy of predicted points or homography. However, it can effectively omit the jittering and provide smoother results.  

\section{Acknowledgement}
This work is supported by Stathletes through the Mitacs Accelerate Program.

\bibliographystyle{IEEEbib}
\bibliography{strings,refs}

\end{document}